\documentclass{article}
\usepackage{amssymb}
\usepackage{graphicx}
\usepackage{url}

\usepackage{mystylefile}

\title{Clustering via Mode Seeking by Direct Estimation of the Gradient
of a Log-Density} 
\author{Hiroaki Sasaki \\ sasaki@sg.cs.titech.ac.jp \\ Graduate School
of Information Science and Engineering, \\ Tokyo Institute of
Technology, Tokyo, Japan \\\vspace{-1mm} \\ 
Aapo Hyv{\"a}rinen \\ aapo.hyvarinen@helsinki.fi \\ Department of
Computer Science and HIIT,\\ University of Helsinki, Helsinki, Finland
\\\vspace{-1mm} \\
Masashi Sugiyama \\ 
sugi@cs.titech.ac.jp \\
Graduate School of Information Science and
Engineering, \\ Tokyo Institute of Technology, Tokyo, Japan}
\date{}
\begin{document} 
\maketitle

\begin{abstract} 
 \emph{Mean shift clustering} finds the \emph{modes} of the data
 probability density by identifying the zero points of the density
 gradient.  Since it does not require to fix the number of clusters in
 advance, the mean shift has been a popular clustering algorithm in
 various application fields.  A typical implementation of the mean shift
 is to first estimate the density by kernel density estimation and then
 compute its gradient.  However, since good density estimation does not
 necessarily imply accurate estimation of the density gradient, such an
 indirect two-step approach is not reliable.  In this paper, we propose
 a method to \emph{directly} estimate the gradient of the log-density
 without going through density estimation.  The proposed method gives
 the global solution analytically and thus is computationally efficient.
 We then develop a mean-shift-like fixed-point algorithm to find the
 modes of the density for clustering. As in the mean shift, one does not
 need to set the number of clusters in advance. We empirically show that
 the proposed clustering method works much better than the mean shift
 especially for high-dimensional data. Experimental results further
 indicate that the proposed method outperforms existing clustering
 methods.
\end{abstract} 

\section{Introduction}
\label{sec:intro}
Seeking the \emph{modes} of a probability density has led to a powerful
clustering algorithm called the \emph{mean
shift}~\cite{cheng1995mean,comaniciu2002mean,fukunaga1975estimation}. In
the mean shift algorithm, all input samples are initially regarded as
candidates of the modes of the density and they are iteratively updated
and merged.  Finally, clustering is performed by associating the input
samples with the obtained modes.  An advantage of the mean shift is that
the number of clusters does not need to be specified in advance.  Thanks
to this extremely useful property, the mean shift has been successfully
employed in various applications such as image
segmentation~\cite{comaniciu2002mean,tao2007color,wang2004image} and
object tracking~\cite{collins2003mean,comaniciu2000real}.

In mode seeking, a central technical challenge is accurate estimation of
the gradient of a density. The mean shift takes a two-step approach:
kernel density estimation (KDE) is first used to approximate the density
and then its gradient is computed.  However, such a two-step approach
performs poorly because a good estimator of the density does not
necessarily mean a good estimator of the density gradient.  In
particular, KDE tends to produce a smooth density estimate and therefore
the density is over-flattened.  This yields the modes in a multi-modal
density to be collapsed.  Furthermore, KDE itself tends to perform
poorly in high-dimensional problems~\cite{comaniciu2002mean}.

To overcome this problem, we propose a method called \emph{least-squares
log-density gradients} (LSLDG), which \emph{directly} estimates the
gradient of a log-density by least-squares without going through density
estimation.  The proposed method can be regarded as a non-parametric
extension of \emph{score matching}
\cite{hyvarinen2005estimation,sriperumbudur2013density}, which has
originally been developed for least-squares parametric density
estimation with intractable partition functions.  We then derive a
fixed-point algorithm to find the modes of the density, which is our
proposed clustering algorithm called \emph{LSLDG clustering}.

All tuning parameters included in LSLDG such as the Gaussian kernel
width and the regularization parameter can be objectively optimized by
cross-validation in terms of the squared error.  Furthermore, since
LSLDG clustering inherits the same algorithmic structure as the original
mean shift, it does not require the number of clusters to be fixed in
advance.  Thus, LSLDG clustering does not involve \emph{any} tuning
parameters to be manually determined, which is a significant advantage
over standard clustering algorithms such as \emph{spectral clustering}
\cite{nips02-AA35}, because clustering is an unsupervised learning
problem and appropriately controlling tuning parameters is generally
very hard. A recent study based on \emph{information-maximization
clustering}~\cite{NC:Sugiyama+etal:2014} provided an
information-theoretic mean to determine tuning parameters objectively,
but it still requires the user to fix the number of clusters in advance.

The remainder of this paper is structured as follows.  We derive a
method to directly estimate the gradient of a log-density in
Section~\ref{sec:dire}, and then use it for finding clusters in the data
in Section~\ref{sec:modeseek}.  Various possibilities for extension are
discussed in Section~\ref{extensions}, and the usefulness of the
proposed method is experimentally investigated in
Section~\ref{sec:artdata}.  Finally this paper is concluded in
Section~\ref{conclusions}.
\section{Direct Estimation of the Gradient of a Log-density}
\label{sec:dire} 
In this section, we propose a method to estimate the log-density
gradient.
\begin{figure}[t]
  \centering \includegraphics[scale=0.7]{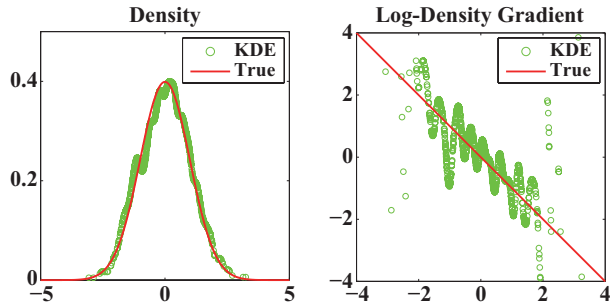}
  \caption{\label{fig:BadKDE} An example of a good density estimation,
  but inaccurate estimation of the log-density gradient by KDE.}
\end{figure} 
\subsection{Problem Formulation}
Let us consider a probability distribution on $\mathbb{R}^d$ with
density $p^*(\vector{x})$, which is unknown but $n$ i.i.d.~samples
$\mathcal{X}=\{\vector{x}_i\}_{i=1}^n$ are available.  Our goal is to
estimate the gradient of the logarithm of the density $p^*(\vector{x})$
with respect to $\vector{x}$ from $\mathcal{X}$:
\begin{align*}
 \vector{g}^*(\vector{x})=(g_1^*(\vector{x}),\ldots,g_d^*(\vector{x}))^\top
  =\nabla \log p^*(\vector{x})
  =\frac{\nabla p^*(\vector{x})}{p^*(\vector{x})}.
\end{align*}
A naive approach to estimate $\vector{g}^*(\vector{x})$ is to first
obtain a density estimate $\widehat{p}(\vector{x})$ and then compute its
log-gradient $\nabla \log \widehat{p}(\vector{x})$.  However, this
two-step approach does not work well because a good density estimate
$\widehat{p}(\vector{x})$ does not necessarily provide an accurate
estimate of its log-gradient $\nabla \log \widehat{p}(\vector{x})$. For
example, in Figure~\ref{fig:BadKDE}, density estimation is performed
fairly well, but its log-density gradient is quite inaccurate. Below, we
describe a method to directly estimate the log-density gradient $\nabla
\log p^*(\vector{x})$ without going through density estimation. The
method is based on the mathematics of score matching
\cite{hyvarinen2005estimation}, the difference being that our goal is
to approximate the gradient of the log-density instead of model
parameter estimation.
\subsection{Least-Squares Log-Density Gradients}
\label{ssec:esti}
Our basic idea is to directly fit a model
$\vector{g}(\vector{x})=(g_1(\vector{x}),\ldots,g_d(\vector{x}))^\top$
to the true log-density gradient $\vector{g}^*(\vector{x})$ under the
squared loss:
\begin{align*}
 J_j(g_j)&=\int
 \Big(g_j(\vector{x})-g_j^*(\vector{x})\Big)^2p^*(\vector{x})
 \mathrm{d}\vector{x}-\int
g_j^*(\vector{x})^2p^*(\vector{x}) \mathrm{d}\vector{x}
\\
&=\int g_j(\vector{x})^2p^*(\vector{x}) \mathrm{d}\vector{x} -2\int
g_j(\vector{x}) g_j^*(\vector{x}) p^*(\vector{x}) \mathrm{d}\vector{x}\\
&=\int g_j(\vector{x})^2p^*(\vector{x}) \mathrm{d}\vector{x} -2\int
g_j(\vector{x})\partial_j p^*(\vector{x}) \mathrm{d}\vector{x}\\ &=\int
g_j(\vector{x})^2p^*(\vector{x}) \mathrm{d}\vector{x} +2\int \partial_j
g_j(\vector{x}) p^*(\vector{x}) \mathrm{d}\vector{x},
\end{align*}
where $\partial_j$ denotes the partial derivative with respect to the
$j$-th variable of $\vector{x}$ and the last equality follows from
\emph{integration by parts} under some
conditions~\cite{hyvarinen2005estimation}.  Then the empirical
approximation of $J_j$ is given as
\begin{align}
 \widehat{J}_j(g_j)=\frac{1}{n}\sum_{i=1}^n g_j(\vector{x}_i)^2
+\frac{2}{n}\sum_{i=1}^n\partial_j g_j(\vector{x}_i). \label{eqn:L2sample}
\end{align}
As the model $g_j(\vector{x})$, we use the following linear-in-parameter
model, which is related to using an exponential family:
 \begin{align*}
 g_j(\vector{x})&=\sum_{i=1}^n \theta_{i,j}\psi_{i,j}(\vector{x})
 =\vector{\theta}_j^\top \vector{\psi}_j(\vector{x}), 
\end{align*}
where $\vector{\theta}$ denotes the parameter vector and
$\psi_{i,j}(\vector{x})$ is a basis function.  This model further yields
\begin{align*}
  \partial_j g_j(\vector{x})&=\sum_{i=1}^n
  \theta_{i,j}\partial_j\psi_{i,j}(\vector{x})
  =\vector{\theta}_j^{\top}\vector{\varphi}_j(\vector{x}),
\end{align*}
where
$\vector{\varphi}_j(\vector{x})=(\partial_j\psi_{1,j}(\vector{x}),\dots,\partial_j\psi_{n,j}(\vector{x}))$.

Adding an $\ell_2$-regularizer to (\ref{eqn:L2sample}), the optimization
problem is compactly expressed as
\begin{align}
\widehat{\vector{\theta}}_j=&\mbox{ arg}\min_{\vector{\theta}_j} \left[
\vector{\theta}_j^{\top}
\vector{G}^{(j)}\vector{\theta}_j+2\vector{\theta}_j^{\top}\vector{h}_j
+\lambda\vector{\theta}_j^{\top}\vector{\theta}_j \right],
 \label{eqn:optpara}
\end{align}
where $\lambda\ge0$ is the regularization parameter, and $\vector{G}^{(j)}$ and
$\vector{h}_j$ are defined by 
\begin{align*}
\vector{G}^{(j)}=\frac{1}{n}\sum_{i=1}^n
\vector{\psi}_j(\vector{x}_i)\vector{\psi}_j(\vector{x}_i)^{\top},
 ~~~\vector{h}_j =\frac{1}{n}\sum_{i=1}^n\vector{\varphi}_{j}(\vector{x}_i).
\end{align*}
As in score matching for an exponential
family~\cite{hyvarinen2007some}, the optimization problem
\eqref{eqn:optpara} can be solved analytically as
\begin{align*}
 \widehat{\vector{\theta}}_j
 =-(\vector{G}^{(j)}+\lambda\vector{I})^{-1}\vector{h}_j,
\end{align*}
where $\vector{I}$ denotes the identity matrix.
Finally, we obtain the estimator $\widehat{g}_j$ as
\begin{align*}
 \widehat{g}_j(\vector{x})=\sum_{i=1}^n\widehat{\theta}_{i,j}\psi_{i,j}(\vector{x})
 =\widehat{\vector{\theta}}_j^{\top}\vector{\psi}_j(\vector{x}).
\end{align*}
We call this method \emph{least-squares log-density gradients} (LSLDG).
\subsection{Model Selection by Cross-Validation}
\label{ssec:model}
The performance of LSLDG depends on the choice of the regularization
parameter $\lambda$ and parameters included in the basis function
$\vector{\psi}_j$.  They can be objectively chosen via cross-validation
as follows:
\begin{enumerate}
 \item Divide the samples $\mathcal{X}=\{\vector{x}_i\}_{i=1}^n$ into $N$ disjoint
       subsets $\{\mathcal{X}_i\}_{i=1}^N$.
\item For $i=1,\ldots,N$
\begin{enumerate}
\item Compute the LSLDG estimator $\widehat{g}^{(i)}_j$ from
      $\mathcal{X}\backslash\mathcal{X}_i$ (i.e.,
  all samples except $\mathcal{X}_i$).
 \item Compute its hold-out error for $\mathcal{X}_i$:
       \begin{align*}
	\text{CV}^{(i)}=
	\frac{1}{|\mathcal{X}_i|}\sum_{\vector{x}\in\mathcal{X}_i}
	\sum_{j=1}^d\left[\widehat{g}^{(i)}_j(\vector{x})^2
	+2\partial_j\widehat{g}^{(i)}_j(\vector{x})\right],
       \end{align*}
       where $|\mathcal{X}_i|$ denotes the cardinality of $\mathcal{X}_i$.
     \end{enumerate}       

 \item Compute the average hold-out error as 
       \begin{align}
	\text{CV}= \frac{1}{N}\sum_{i=1}^N\text{CV}^{(i)}.
	\label{eqn:CV}
       \end{align}

 \item Choose the model that minimizes (\ref{eqn:CV}) with respect to
       $\lambda$ and parameters in $\vector{\psi}_j$, and compute the
       final LSLDG estimator $\widehat{g}_j$ with the chosen model using
       all samples $\mathcal{X}$ .
\end{enumerate}

\section{Clustering via Mode Seeking}
\label{sec:modeseek}
In this section, we derive a clustering algorithm based on LSLDG.  Our
basic idea follows the same line as the \emph{mean shift} algorithm
\cite{cheng1995mean,comaniciu2002mean,fukunaga1975estimation}, i.e., to
assign each data sample to a nearby \emph{mode} of the density.

\subsection{Gradient-Based Approaches}

A naive implementation of this idea is to use \emph{gradient ascent} for
each data sample to let it converge to one of the modes of the density
in the vicinity:
\begin{align*}
  \vector{x}_i\longleftarrow
  \vector{x}_i+\varepsilon\widehat{\vector{g}}(\vector{x}_i),
\end{align*}
where $\varepsilon>0$ is the step size. Since
\begin{align*}
  \nabla \log p(\vector{x})=\frac{\nabla
  p(\vector{x})}{p(\vector{x})}\propto\nabla p(\vector{x}),
\end{align*}
the gradient of the log-density $\log p(\vector{x})$ keeps the same
direction as the gradient of the original density $p(\vector{x})$.
However, due to $p(\vector{x})$ in the denominator, the log-gradient
vector gets longer when $p(\vector{x})<1$ and shorter when
$p(\vector{x})>1$.  This is practically a suitable adjustment because
$p(\vector{x})<1$ ($p(\vector{x})>1$) often means that the current point
$\vector{x}$ is far from (close to) a mode. Indeed, The faster
convergence of gradient ascent with the log-density was justified in the
same way~\cite{fukunaga1975estimation}. To further increase the speed of
convergence, using a \emph{quasi-Newton} method is also promising:
\begin{align*}
  \vector{x}_i\longleftarrow
  \vector{x}_i+\varepsilon\widehat{\vector{Q}}\widehat{\vector{g}}(\vector{x}_i),
\end{align*}
where $\widehat{\vector{Q}}$ is an estimate of the inverse Hessian matrix.

\subsection{Fixed-Point Approach}
\label{ssec:fixed-point} 
In the gradient-based approach, choosing the step size parameter
$\varepsilon$ is a crucial problem. To avoid this problem, we develop a
\emph{fixed-point method}, in analogy to the original mean-shift
method. To easily derive a fixed-point equation, we focus on the basis
function of the following form:
\begin{align*}
  \psi_{i,j}(\vector{x})=
 \frac{1}{\sigma^2}[\vector{c}_{i}-\vector{x}]_j\phi_i(\vector{x}),
\end{align*}
where $\sigma^2$ is a constant, $\vector{c}_{i}$ is a $d$-dimensional
constant vector, $\phi_i(\vector{x})$ is a ``mother'' basis function,
and $[\cdot]_j$ denotes the $j$-th element of a vector.  A typical
choice of the mother basis function $\phi_i(\vector{x})$ is the Gaussian
function:
\begin{align}
 \phi_i(\vector{x})=\exp
 \left(-\frac{\|\vector{x}-\vector{c}_{i}\|^2}{2\sigma^2}\right),
 \label{phi=Gaussian}
\end{align}
where the Gaussian center $\vector{c}_{i}$ may be fixed at sample
$\vector{x}_{i}$. In experiments, we only use $100$ Gaussian centers
chosen randomly from $\{\vector{x}_i\}_{i=1}^n$. This reduction of
Gaussian centers significantly decreases the computation costs without
sacrificing the performance, as shown in Section 5.2.

For this model, the LSLDG solution can be expressed as
\begin{align*}
  \widehat{g}_j(\vector{x})&=\sum_{i=1}^n
  \widehat{\theta}_{i,j}\psi_{i,j}(\vector{x})
  =\frac{1}{\sigma^2}\sum_{i=1}^n\widehat{\theta}_{i,j}[\vector{c}_{i}-\vector{x}]_j\phi_i(\vector{x})\\
  &=\frac{1}{\sigma^2}\sum_{i=1}^n\widehat{\theta}_{i,j}\phi_i(\vector{x})[\vector{c}_{i}]_j
  -\frac{[\vector{x}]_j}{\sigma^2}\sum_{i=1}^n\widehat{\theta}_{i,j}\phi_i(\vector{x}).
\end{align*}
If $\sum_{i=1}^n\widehat{\theta}_{i,j}\phi_i(\vector{x})\neq0$, setting
$\widehat{g}_j(\vector{x})$ to zero yields
\begin{align}
  [\vector{x}]_j=
  \frac{\sum_{i=1}^n\widehat{\theta}_{i,j}\phi_i(\vector{x})[\vector{c}_{i}]_j}
  {\sum_{i=1}^n\widehat{\theta}_{i,j}\phi_i(\vector{x})}.
  \label{fixed-point-xj}
\end{align}
We propose to use this equation as a fixed-point update formula by
iteratively substituting the right-hand side to the left-hand side. In
the vector-matrix form, the update formula is compactly expressed as
\begin{align*}
  \vector{x}_i\longleftarrow \vector{B}\vector{\phi}(\vector{x}_i)
  ./(\widehat{\vector{\Theta}}^\top\vector{\phi}(\vector{x}_i)),
\end{align*}
where $B_{j,i}=\widehat{\theta}_{i,j}[\vector{c}_i]_j$,
$\widehat{\Theta}_{i,j}=\widehat{\theta}_{i,j}$,
$\vector{\phi}(\vector{x})=(\phi_1(\vector{x}),\ldots,\phi_n(\vector{x}))^\top$,
and ``$./$'' denotes the element-wise division.

This update formula is similar to the one used in the mean shift
algorithm \cite[Eq.(20)]{comaniciu2002mean}, which corresponds to
$\widehat{\theta}_{i,j}=1/n$:
\begin{align*}
  \vector{x}\longleftarrow
  \frac{\sum_{i=1}^n\phi_i(\vector{x})\vector{c}_{i}}
  {\sum_{i=1}^n\phi_i(\vector{x})},
\end{align*}
where $\phi_i$ is typically chosen as the Gaussian function
\eqref{phi=Gaussian}.  Thus, the proposed method can be regarded as a
weighted variant of the mean shift algorithm, where the weights
$\widehat{\theta}_{i,j}$ are learned by LSLDG.  A similar weighted mean
shift method has already been studied in \cite{cheng1995mean}, but the
weights were determined heuristically.

Cheng proved that the mean shift update is equivalent to a gradient
ascent with an adaptive step size~\cite{cheng1995mean}. LSLDG clustering
also inherits this property. If
$[\vector{x}]_j\sum_{i=1}^n\widehat{\theta}_{i,j}\phi_i(\vector{x})$ is
subtracted from and added to the numerator of Eq.\eqref{fixed-point-xj}
(thus the equation remains the same), we obtain
\begin{align*}
  [\vector{x}]_j=[\vector{x}]_j+\varepsilon_j(\vector{x})\widehat{g}_j(\vector{x}),
\end{align*}
where
\begin{align*}
  \varepsilon_j(\vector{x})=
  \frac{\sigma^2}{\sum_{i=1}^n\widehat{\theta}_{i,j}\phi_i(\vector{x})}.
\end{align*}
This shows that our fixed-point update rule can be regarded as a
gradient ascent with an adaptive step size $\varepsilon_j(\vector{x})$.

If $\phi_i(\vector{x})$ is set to be the Gaussian function
\eqref{phi=Gaussian},
$\sum_{i=1}^n\widehat{\theta}_{i,j}\phi_i(\vector{x})$ can actually be
regarded as an estimate of the original log-density $\log
p^*(\vector{x})$.  More specifically, we can easily see that the partial
derivative of $\phi_i(\vector{x})$ with respect to the $j$-th variable
of $\vector{x}$ is $\psi_{i,j}(\vector{x})$:
\begin{align*}
\partial_j \phi_i(\vector{x})=\psi_{i,j}(\vector{x}).
\end{align*}
Then we have
\begin{align*}
\partial_j \log p^*(\vector{x}) &=g^*_j(\vector{x}) \approx
\widehat{g}_j(\vector{x})
=\sum_{i=1}^n\widehat{\theta}_{i,j}\psi_{i,j}(\vector{x})\\
&=\sum_{i=1}^n\widehat{\theta}_{i,j}\partial_j \phi_i(\vector{x})
=\partial_j\sum_{i=1}^n\widehat{\theta}_{i,j} \phi_i(\vector{x}).
\end{align*}
This implies that $\sum_{i=1}^n\widehat{\theta}_{i,j}
\phi_i(\vector{x})$ is an estimate of $\log p^*(\vector{x})$ up to a
constant.  Therefore, when $\log p^*(\vector{x})$ is small (large), the
proposed fixed-point algorithm increases (decreases) the adaptive step
size $\varepsilon_j(\vector{x})$ to more aggressively (conservatively)
ascend the gradient.  This step-size adaptation would be reasonable
because small (large) $\log p^*(\vector{x})$ often means that the
current solution is far from (close to) a mode.
\section{Extensions}
\label{extensions} 
In the previous section, we focused on the simplest
setting to clearly convey the essence of the proposed idea.  However, we
can easily extend the proposed method to various directions.  In this
section, we discuss such possibilities.

\subsection{Common Basis Functions}
When the basis function is common to all dimensions, i.e.,
$\vector{\psi}_j(\vector{x})=\vector{\psi}(\vector{x})$ for
$j=1,\ldots,d$, the matrix $\vector{G}^{(j)}$ becomes independent of $j$
as
$\vector{G}=\frac{1}{n}\sum_{i=1}^n\vector{\psi}(\vector{x}_i)\vector{\psi}(\vector{x}_i)^{\top}$.
Then, matrix inverse has to be computed only once for all dimensions:
\begin{align*}
  (\widehat{\vector{\theta}}_1,\ldots,\widehat{\vector{\theta}}_d)
 =-(\vector{G}+\lambda\vector{I})^{-1}(\vector{h}_1,\ldots,\vector{h}_d).
\end{align*}
This significantly speeds up the computation 
particularly when the dimensionality $d$ is high.

\subsection{Multi-Task Learning}
The above common-basis setup allows us to employ the \emph{regularized multi-task} method \cite{KDD:Evgeniou+Pontil:2004},
by regarding the estimation problem of $g_j^*(\vector{x})$ as the $j$-th task.
The basic idea of regularized multi-task learning
is that, if $g_j^*(\vector{x})$ and $g_{j'}^*(\vector{x})$ are similar to each other,
the corresponding parameters $\vector{\theta}_j$ and $\vector{\theta}_{j'}$ are imposed to be close to each other.
This idea can be implemented in the regularization framework as
\begin{align*}
&\min_{\vector{\theta}_1,\ldots,\vector{\theta}_d} \left[
\sum_{j=1}^d\left(\vector{\theta}_j^{\top}
\vector{G}^{(j)}\vector{\theta}_j+2\vector{\theta}_j^{\top}\vector{h}_j
+\lambda_j\vector{\theta}_j^{\top}\vector{\theta}_j\right)+\gamma\sum_{j,j'=1}^d\gamma_{j,j'}\|\vector{\theta}_j-\vector{\theta}_{j'}\|^2\right],
\end{align*}
where $\lambda_j>0$ is the ordinary regularization parameter for the $j$-th task,
$0\le\gamma_{j,j'}\le1$ is the similarity between the $j$-th task and the $j'$-th task,
and $\gamma>0$ controls the strength of this multi-task regularizer.
A notable advantage of this regularization approach is that the solution can be obtained analytically.
When the task similarity $\gamma_{j,j'}$ is unknown, task similarity and solutions may be iteratively learned.
More specifically, starting from $\gamma_{j,j'}=1$ for all $j,j'=1,\ldots,d$,
the solutions $\vector{\theta}_1,\ldots,\vector{\theta}_d$ are computed.
Then, task similarity is updated, e.g., 
by $\gamma_{j,j'}=\exp(-\|\vector{\theta}_j-\vector{\theta}_{j'}\|^2)$ for $j,j'=1,\ldots,d$,
and the solutions $\vector{\theta}_1,\ldots,\vector{\theta}_d$ are computed again.

Another multi-task idea called \emph{multi-task feature learning}
\cite{NIPS2006_251} may also be applied in log-density gradient
estimation, which finds the features that can be shared by multiple
tasks.

\subsection{Sparse Estimation}
Instead of the $\ell_2$-regularizer $\lambda\|\vector{\theta}_j\|^2$,
the $\ell_1$-regularizer $\lambda\|\vector{\theta}_j\|_1$ may be used to
obtain a sparse solution \cite{JRSS:Tibshirani:1996}, which can be
computed more efficiently.  The entire regularization path (i.e., the
solutions for all $\lambda\ge0$) can also be computed efficiently, based
on the piece-wise linearity of the solution path with respect to
$\lambda$ \cite{JMLR:Hastie+etal:2004}.

\subsection{Bregman Loss}
The squared loss can be generalized to the \emph{Bregman
loss}~\cite{USSR-CMMP:Bregman:1967}. More specifically, for $f$ being a
differentiable and strictly convex function and $C^{(f)}_j=\int
f(g_j^\ast(\vector{x})) p^*(\vector{x}) \mathrm{d}\vector{x}$,
\begin{align*}
  J^{(f)}_j(g_j)&=
  \int \left(
 f(g_j^\ast(\vector{x}))-f(g_j(\vector{x}))-f'(g_j(\vector{x}))
 (g_j^\ast(\vector{x})-g_j(\vector{x}))\right)
  p^*(\vector{x}) \mathrm{d}\vector{x}-C^{(f)}_j\\
  &=
  \int \left(-f(g_j(\vector{x}))+f'(g_j(\vector{x}))g_j(\vector{x})\right)
  p^*(\vector{x}) \mathrm{d}\vector{x}
 -\int f'(g_j(\vector{x}))\partial_j p^*(\vector{x}) \mathrm{d}\vector{x}\\
  &=
  \int \left(-f(g_j(\vector{x}))+f'(g_j(\vector{x}))g_j(\vector{x})
 + \partial_j f'(g_j(\vector{x}))\right)p^*(\vector{x}) \mathrm{d}\vector{x},
\end{align*}
where $f'(t)$ is the derivative of $f(t)$ with respect to $t$
and the last equality follows again from integration by parts.
The empirical approximation of $J^{(f)}_j$ is given as
\begin{align*}
 \widehat{J}^{(f)}_j(g_j)&=\frac{1}{n}\sum_{i=1}^n
\left(-f(g_j(\vector{x}_i))+f'(g_j(\vector{x}_i))g_j(\vector{x}_i)
+ \partial_j f'(g_j(\vector{x}_i))\right).
\end{align*}
When $f(t)=t^2$, the Bregman loss is reduced to the squared loss and we
can recover the LSLDG criterion \eqref{eqn:L2sample}.  On the other
hand, $f(t)=-\log t$ gives the \emph{Kullback-Leibler
loss}~\cite{Annals-Math-Stat:Kullback+Leibler:1951}, $f(t)=t\log
t-(1+t)\log(1+t)$ gives the \emph{logistic loss}
\cite{AISM:Sugiyama+etal:2012}, and $f(t)=(t^{1+\alpha}-t)/\alpha$ for
$\alpha>0$ gives the \emph{power loss} \cite{Biometrika:Basu+etal:1998}.
Although each choice has its own specialty, e.g., the power loss
possesses high robustness against outliers, the squared loss was shown
to be endowed with the highest numerical stability in terms of the
\emph{condition number} \cite{ML:Kanamori+etal:2013}.

\subsection{Blurring Mean Shift}
Fukunaga and Hostetler originally proposed a mean shift algorithm for
updating not only the data points but also the density estimation at
each iteration~\cite{fukunaga1975estimation}. Later, this algorithm was
named the \emph{blurring mean
shift}~\cite{carreira2006fast,cheng1995mean}. Combined with
the idea of the blurring mean shift, another possible algorithm for
LSLDG clustering is to re-estimate the log-density gradient at each
iteration for new data points. This algorithm hopefully works well as
the blurring mean shift does~\cite{carreira2006fast}.
\section{Experiments}
\label{sec:artdata}
\begin{figure}[t!]
  \centering \includegraphics[scale=0.5]{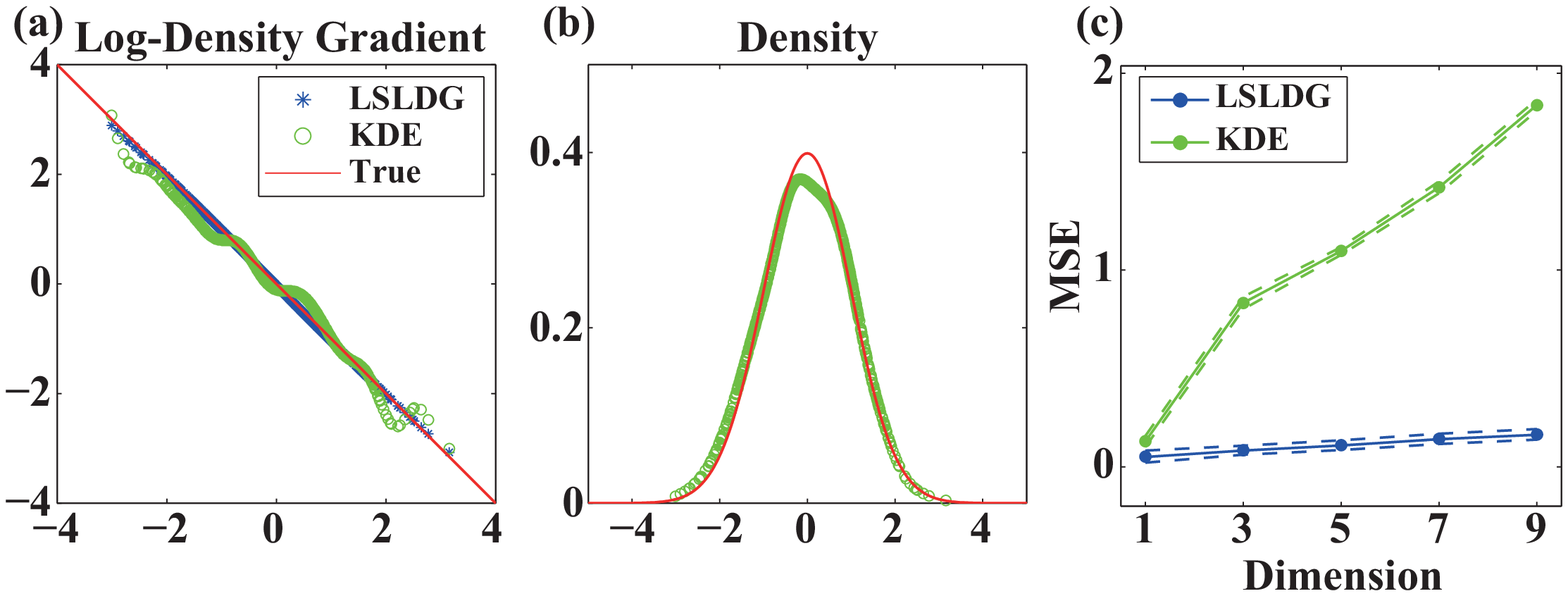}
  \caption{\label{fig:KDEvsGau} LSLDG vs. KDE for Gaussian data. (a)
  Profiles of the true log-density gradient and its estimates obtained
  by LSLDG and KDE.  (b) True and estimated densities by KDE. (c)
  Averages and standard deviations of mean-squared errors to the true
  log-density gradients as functions of input dimensionality over $100$
  runs.}
 \vspace*{3mm}
 \includegraphics[scale=0.5]{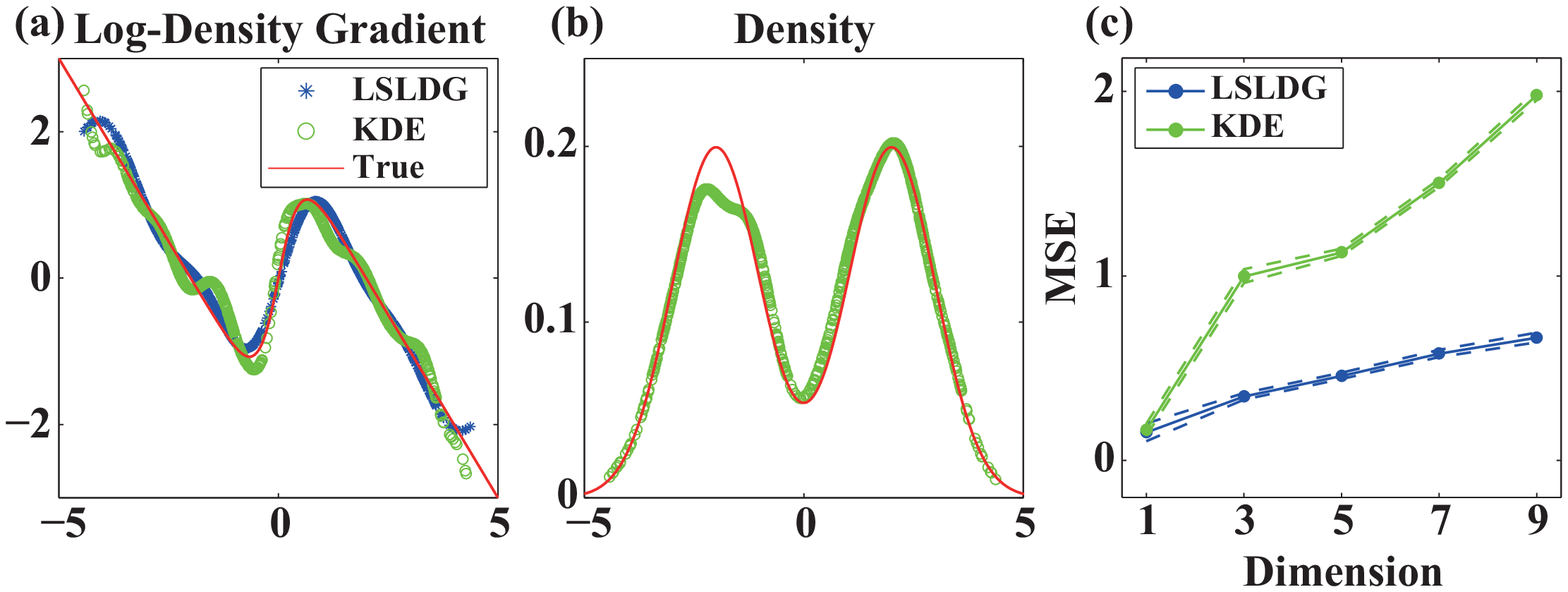}
   \caption{\label{fig:KDEvsGMM} LSLDG vs. KDE for data sampled from a
 mixture of two Gaussians.}
\end{figure} 
In this section, we demonstrate the usefulness of the proposed LSLDG method.
\subsection{Illustration of Log-Density Gradient Estimation}
\label{ssec:accuracy}
We first illustrate how LSLDG estimates log-density gradients
using $n=1,000$ samples drawn from $p(\vector{x})$ where either
\begin{itemize}
\item $p(\vector{x})$ is the standard normal density, or
\item $p(\vector{x})$ is a mixture of two Gaussians with
  means $2$ and $-2$, variances $1$ and $1$, and mixing coefficients $0.5$
  and $0.5$. 
\end{itemize}
As described in Section~\ref{ssec:model}, the Gaussian width $\sigma$ and the
regularization parameter $\lambda$ are chosen by 5-fold
cross-validation from the following candidate set:
\begin{align}
\{10^{-2},10^{-1.5},10^{-1},10^{-0.5},10^{0},10^{0.5},10^{1}\}.
 \label{eqn:candidates}
\end{align}
We compare the performance of the proposed method with Gaussian KDE,
where the Gaussian width is chosen by likelihood cross-validation from
the same candidate set in (\ref{eqn:candidates}).

The results for the Gaussian data are presented in
Figure~\ref{fig:KDEvsGau}. This shows that LSLDG gives a nice smooth
estimate, while the estimate by KDE is rather oscillating.  Note that
KDE still works well as a density estimator (see
Figure~\ref{fig:KDEvsGau}(b)).  This clearly illustrates that a good
density estimate (obtained by KDE) does not necessarily yield a good
estimate of the log-density gradient.  We repeated this experiment $100$
times and the mean squared error to the true log-density gradient is
plotted in Figure~\ref{fig:KDEvsGau}(c) as a function of the input
dimensionality.  This shows that while the error of KDE increases
sharply as a function of dimensionality, that of LSLDG increases only
mildly.  This implies that the advantage of directly estimating the
log-density gradient is more prominent in high-dimensional cases.
Similar tendencies can be observed also for the Gaussian mixture data
(Figure~\ref{fig:KDEvsGMM}), where the added dimensions in
Figure~\ref{fig:KDEvsGMM}(c) simply follow the standard normal
distribution.
\subsection{Illustration of Clustering}
\label{ssec:clustering}
\begin{figure}[t!]
  \centering \includegraphics[scale=0.63]{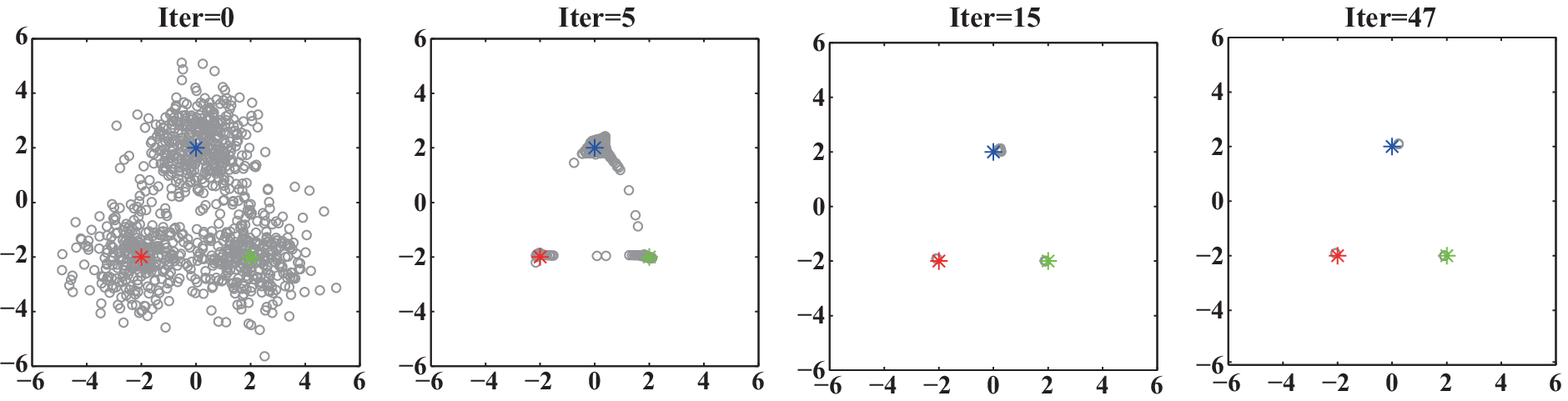}
   \caption{\label{fig:GMMClusters} Transition of data points toward the
   modes at the initial state, $5$th, $15$th, and $47$th iterations.
   The blue, red, and green symbols represent the three centers of the
   Gaussian mixture model.}
 \centering
 \vspace*{3mm}
 \includegraphics[scale=0.6]{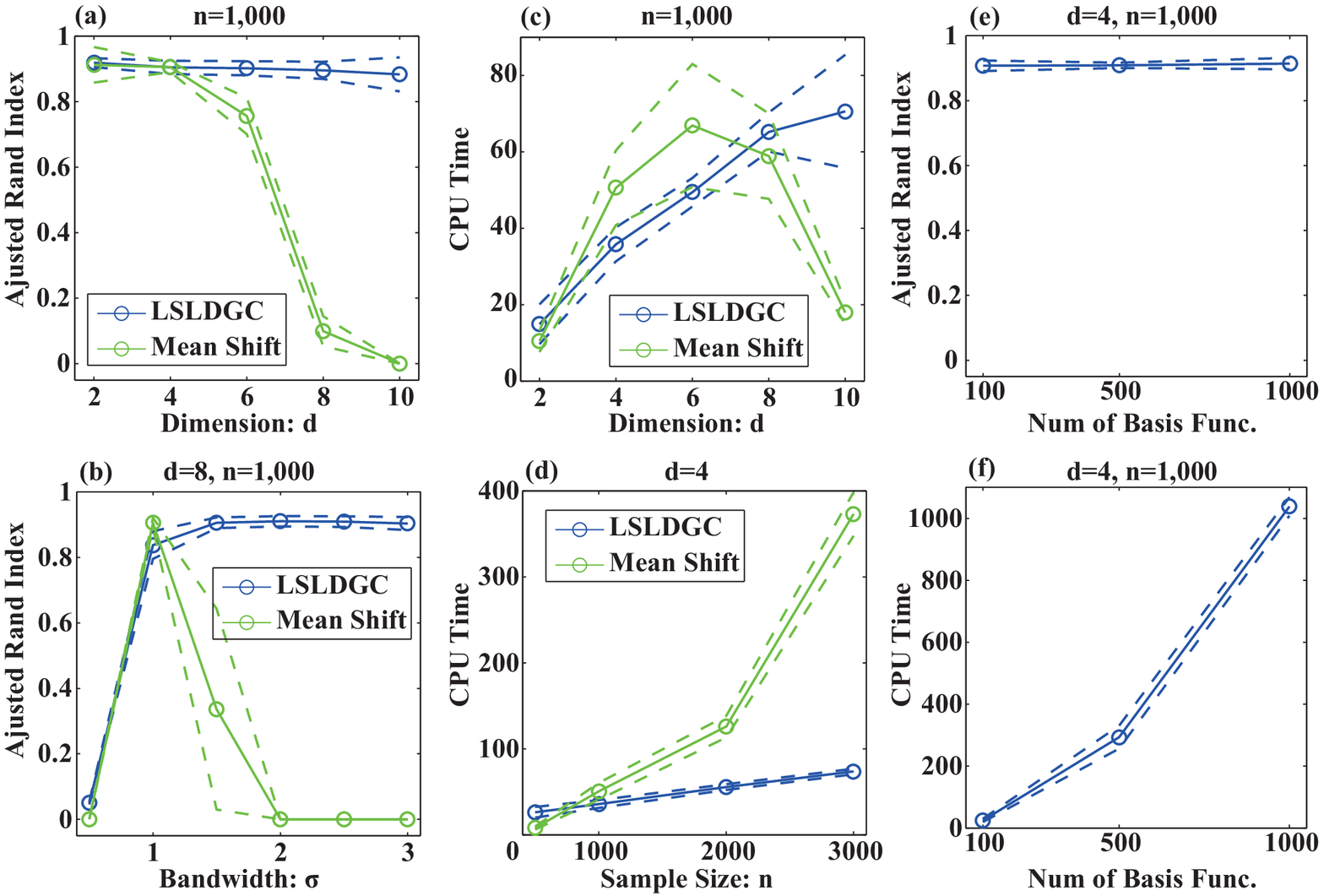}
 \caption{\label{fig:ClustNum} Means and standard deviations of
 clustering performance over $100$ runs measured by ARI as functions of
 (a) dimensionality of data and (b) the Gaussian width (when
 dimensionality is $8$). CPU time is also compared with respect to (c)
 dimensionality and (d) sample size. (e) ARI and (f) CPU time for LSLDG
 clustering are plotted as functions of the number of basis functions.}
  \end{figure} 
Next, we illustrate the behavior of LSLDG clustering on $1,000$ samples
gathered from the mixture of three Gaussians whose means were $(0,2)$,
$(-2,-2)$, and $(2,-2)$, and covariance matrices were the identity
matrices. The mixing coefficients were $0.4$, $0.3$, and $0.3$.
Figure~\ref{fig:GMMClusters} illustrates the transition of data
samples over update iterations, showing that all points converge to the
nearest modes within 47 iterations.

We compare the performance of the proposed method with the Gaussian mean
shift~\cite{carreira2007gaussian,cheng1995mean}. To investigate the
effect of high dimensionality, further dimensions following the standard
normal distribution are added to data points.  We measure the clustering
performance by the \emph{adjusted Rand index} (ARI)
\cite{JoC:Hubert+Arabie:1985}, which takes the maximum value $1$ when
clustering is perfect.

ARI values are plotted as a function of input dimensionality in
Figure~\ref{fig:ClustNum}(a) averaged over $100$ runs.  When the
dimensionality of data is in the range of $2$--$4$, both methods work
very well. However, when the dimensionality is beyond $4$, the
performance of the Gaussian mean shift drops sharply. In contrast, for
the proposed method, reasonably high ARI values are still attained even
when the dimensionality is increased.

Figure~\ref{fig:ClustNum}(b) plots the ARI values for $d=8$ when the
Gaussian widths are changed. This shows that the proposed LSLDG
clustering performs well for a wide range of Gaussian widths, while the
ARI plot for the Gaussian mean shift is peaky.  This implies that
selection of Gaussian widths is much harder for the Gaussian mean shift
than LSLDG clustering.

LSLDG clustering is also advantageous in terms of the computational
costs. Figure~\ref{fig:ClustNum}(c) shows that CPU time of LSLDG
clustering is almost the same as or shorter than that of the mean shift,
when the ARI values for both methods are high enough. The shorter CPU
time of the mean shift when the dimensionality is more than 8 comes from
the fact that a smaller bandwidth is chosen; then the number of clusters
is more close to the number of kernels and thus the mean shift converges
very quickly, although this performs poorly. With the same sample size,
LSLDG clustering is much faster than the mean shift, as plotted in
Figure~\ref{fig:ClustNum}(d). The speedup was brought by reducing the
kernel centers, which was shown to significantly improve the computation
costs without worsening the clustering performance, as depicted in
Figures~\ref{fig:ClustNum}(e) and (f).
\subsection{Image Discontinuity Preserving Smoothing and Image
  Segmentation} \label{ssec:segmentation}
\begin{figure}[t!]
  \centering \includegraphics[scale=0.7]{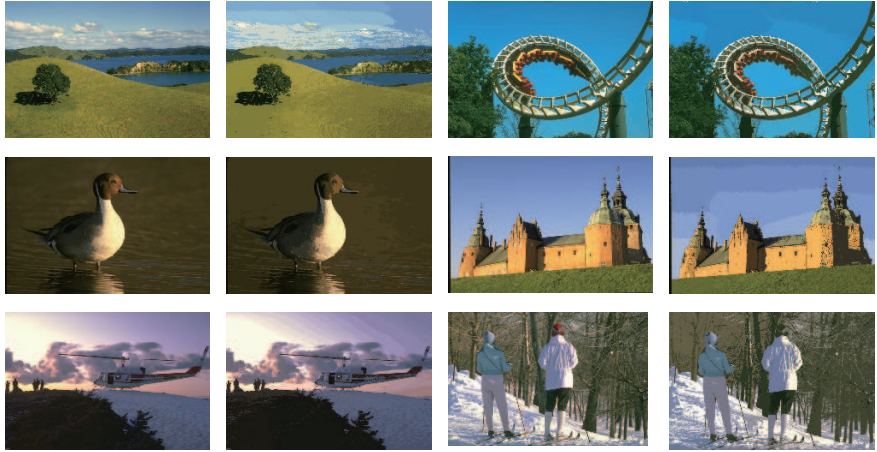}
  \caption{\label{fig:segment} Examples of images after LSLDG
  clustering. The left-hand figure in each pair is the input image, and
  the right-hand one is the image after LSLDG clustering.}
\end{figure}
The mean shift has been successively applied to image discontinuity
preserving smoothing and segmentation
tasks~\cite{comaniciu2002mean,tao2007color,wang2004image}. Here, we
investigate the performance of LSLDG clustering in those tasks.

As image data, we used the \emph{Berkeley segmentation dataset}
(BSD500)~\cite{arbelaez2011contour}.\footnote{\url{http://www.eecs.berkeley.edu/Research/Projects/CS/vision/grouping/resources.html}}
From one image, the information of color (three dimensions) and spatial
positions (two dimensions) were extracted per pixel. Thus, the
dimensionality of data is five, and the total number of samples is the
same as the total number of pixels. As often assumed in the mean
shift~\cite{comaniciu2002mean}, for image data, we used the following
mother basis function:
\begin{align}
 \phi_i(\vector{x})&=
 \exp\left(-\frac{\|\vector{x}^{c}-\vector{c}_i^c\|}{2\sigma_c^2}\right)
 \exp\left(-\frac{\|\vector{x}^{s}-\vector{c}_i^s\|}{2\sigma_s^2}\right),
\label{eqn:IMbasis}
\end{align}
$\vector{x}^c$ and $\vector{x}^s$ denote the elements for colors and
spatial positions in a data vector $\vector{x}$,
respectively. $\vector{c}_i^c$ and $\vector{c}_i^s$ are the Gaussian
centers. For the two Gaussian width $\sigma_c$ and $\sigma_s$,
cross-validation was performed as in Section~\ref{ssec:model}. In this
experiment, we used a reduced image ($11$ by $16$ or $16$ by $11$
pixels) as the Gaussian centers. For the Gaussian mean shift,
(\ref{eqn:IMbasis}) was employed as a Gaussian kernel, and the two
Gaussian widths were cross-validated based on the likelihood.

Six examples of color images after LSLDG clustering are shown in
Figure~\ref{fig:segment}. In the results, some of the segments, such as
grass, are cleanly smoothed out, while the edges outlining the objects
are preserved. These properties are similar to the results for the mean
shift~\cite{comaniciu2002mean}.

Next, to clarify the difference from the mean shift, we compared the
performance measured by ARI. In this experiment, the input images were
reduced to $81$ by $121$ (or $121$ by $81$) pixels. Since this benchmark
dataset contains several ground truths per image, we simply computed the
mean ARI value to all the ground truths.

The ARI values are summarized in Table~\ref{tab:segmentation}. LSLDG
clustering shows a better ARI value on image segmentation.
\begin{table}[t]
\begin{center}
 \caption{\label{tab:segmentation} Mean ARI values for $200$ images. The
 numbers in the parentheses are standard deviations. The difference
 between the methods is statistically significant at level $1\%$ by the
 t-test.}
\begin{tabular}{|c|c|}
\multicolumn{2}{c}{}  \\
 Mean Shift & LSLDGC  \\
 \hline
 0.08(0.03) & {\bf 0.13(0.06)} \\ 
\end{tabular}
\end{center}
\end{table}
\subsection{Performance Comparison to Existing Clustering Methods}
\label{ssec:comparison}
Finally, we compare LSLDG clustering to existing clustering methods
using accelerometric sensor and speech data.

For comparison, we employed K-means
(KM)~\cite{BerkeleySymp:MacQueen:1967}, spectral clustering
(SC)~\cite{IEEE-PAMI:Shi+Malik:2000,nips02-AA35} with the Gaussian
similarity, and Gaussian mean shift. Since the user has to set the
number of clusters in advance for KM and SC, we set it as the true
number of clusters in each dataset. For the Gaussian mean shift, the
Gaussian width was chosen by likelihood cross-validation. For LSLDG, in
this experiment, we modify the linear-in-parameter model as
\begin{align*}
 g_j(\vector{x})= \sum_{i=1}^n\theta_{i}\psi_{i,j}(\vector{x})
 =\vector{\theta}^{\top}\vector{\psi}_j(\vector{x}).
\end{align*}
The main difference from the model introduced in Section~\ref{ssec:esti}
is that the coefficients $\theta_{i}$ do not depend on $j$, namely, the
dimensionality of data. This modification considerably decreases the
computational costs to much higher dimensional data.

In this experiment, we used the following two datasets where $d$ denotes
the dimensionality of data, $n$ denotes the number of samples, and $c$
denotes the number of true clusters:
\begin{enumerate}
 \item {\it Accelerometry} $(d=5, n=300,~\mathrm{and}~c=3)$.  The
       \emph{ALKAN}
       dataset\footnote{\url{http://alkan.mns.kyutech.ac.jp/web/data.html}},
       which contains $3$-axis (i.e., x-, y-, and z-axes) accelerometric
       data.

\item {\it Speech} $(d=50, n=400,~\mathrm{and}~c=2)$. An in-house speech
      dataset, which contains short utterance samples recorded from $2$
      male subjects speaking in French with sampling rate $44.1$kHz.
\end{enumerate}
The details of the two datasets can be seen
in~\cite{NC:Sugiyama+etal:2014}. For each dataset, as preprocessing, the
variance was normalized after centering in the element-wise manner.

The experimental results are described in Table~\ref{tab:ARI}. For the
accelerometry dataset, LSLDG clustering shows the best performance of
all the methods in the table. In addition to the superior performance,
another advantage is that LSLDG clustering does not include any
parameters which have to be manually tuned. On the other hand, KM and SC
require the users to fix the number of clusters beforehand, which
largely influences the clustering performance. Thus, LSLDG clustering
would be easier to use in practice. For the speech dataset, LSLDG
outperforms the existing clustering methods again
(Table~\ref{tab:ARI}). Since the dimensionality of the dataset, $d=50$,
is much higher than the accelerometry dataset ($d=5$), LSLDG seems to
perform well on high dimensional data, while the mean shift does not
work well on high-dimensional data, as already indicated in
Section~\ref{ssec:clustering}.
\begin{table}[t]
 \caption{\label{tab:ARI} Mean ARI for various methods over $100$
 runs. The standard deviations are indicated in the parentheses. The
 best method in terms of the average ARI and methods judged to be
 comparable to the best one by the t-test at the significance level
 $1\%$ are described in boldface.}
\begin{center}
\begin{tabular}{|c|c|c|c|}
 \multicolumn{4}{c}{}\\ 
 \multicolumn{4}{c}{Accelerometry ($d=5,~n=300,~\mathrm{and}~c=3$)}\\ 
 KM & SC & Mean Shift & LSLDGC \\ 
 \hline 
 0.50(0.03) & 0.20(0.26) & 0.51(0.05) & {\bf 0.61(0.13)}\\ 
 \multicolumn{4}{c}{}\\ 
 \multicolumn{4}{c}{Speech ($d=50,~n=400,~\mathrm{and}~c=2$)}\\ 
 KM & SC & Mean Shift & LSLDGC \\ 
 \hline 
 0.00(0.00) & 0.00(0.00) & 0.00(0.00) & {\bf 0.13(0.02)}\\ 
\end{tabular}
\end{center}
\end{table}
\section{Conclusions}
\label{conclusions}
In this paper, we developed a method to directly estimate the
log-density gradient, and constructed a clustering algorithm on it.  The
proposed log-density gradient estimator can be regarded as a
non-parametric extension of score
matching~\cite{hyvarinen2005estimation,sriperumbudur2013density}, and
the proposed clustering algorithm can be regarded as an extension of the
mean shift
algorithm~\cite{cheng1995mean,comaniciu2002mean,fukunaga1975estimation}.
The key advantage compared to the mean shift is that the proposed
clustering method works well on high-dimensional data for which the mean
shift works poorly. Furthermore, we showed empirically that the proposed
method outperforms existing clustering methods, even in
higher-dimensional problems.
\bibliography{papers,mybib,sugiyama}
\bibliographystyle{plain}
\end{document}